\documentclass[letterpaper]{article} % DO NOT CHANGE THIS
\usepackage{aaai2026}  % DO NOT CHANGE THIS
\nocopyright
\usepackage{times}  % DO NOT CHANGE THIS
\usepackage{helvet}  % DO NOT CHANGE THIS
\usepackage{courier}  % DO NOT CHANGE THIS
\usepackage[hyphens]{url}  % DO NOT CHANGE THIS
\usepackage{graphicx} % DO NOT CHANGE THIS
\usepackage{natbib}  % DO NOT CHANGE THIS AND DO NOT ADD ANY OPTIONS TO IT
\usepackage{caption} % DO NOT CHANGE THIS AND DO NOT ADD ANY OPTIONS TO IT
\usepackage{amsmath}
\title{Exposing Weaknesses in Emotion
Recognition in Conversations}
\author{
Amir Ben Khalifa\textsuperscript{\rm 1},
Fanny Bezançon \textsuperscript{\rm 2},
Bessam Abdulrazak\textsuperscript{\rm 1},
Amine Trabelsi\textsuperscript{\rm 1}
}

\affiliations{
\textsuperscript{\rm 1}Department of Computer Science, Université de Sherbrooke
Sherbrooke, Quebec, Canada\\
\textsuperscript{\rm 2} Polytech Nantes engineering school, Computer Sciences, Université de Nantes
Nantes, France\\
Amir.ben.khalifa@Usherbrooke.ca
fanny1.bezancon@free.fr
Bessam.Abdulrazak@USherbrooke.ca
Amine.Trabelsi@USherbrooke.ca}

\begin{document}

\maketitle

\begin{abstract}
Emotion Recognition in Conversations (ERC) aims to identify speakers' emotions in multi-turn dialogue. Accurate emotion recognition can support a wide range of applications including empathetic conversational agents, mental health support and educational technologies. While many recent approaches rely on task-specific fine-tuning, such models may exploit dataset-specific cues. A central yet rarely questioned assumption in ERC is that each utterance can be assigned a single unambiguous emotion label. To investigate this, we study ERC using Large Language Models (LLMs) in a zero-shot setting, incorporating preceding conversational turns as context. We show that aggregate metrics mask systematic failures, with errors concentrating around utterances containing negations, exclamations, and interjections, a pattern that is consistent across all models, suggesting benchmark limitations rather than model-specific weaknesses. A controlled re-annotation study with four human annotators confirms this: strong agreement is observed in only 35\% of cases, with neutral utterances dominating high-agreement instances, while a substantial portion of emotional categories fall in low-agreement regimes. These findings suggest that many apparent model errors reflect genuine annotation ambiguity rather than poor emotion understanding. Standard single-label evaluation is therefore insufficient. To address this, we introduce an LLM-as-Judge framework that queries each emotion independently, evaluating plausibility given the conversational context, rather than enforcing a single-label decision.
\end{abstract}

\section{Introduction}
Emotion Recognition in Conversations (ERC) is the task of predicting the emotion of each utterance
in multi-turn dialogues \citep{majumder2019dialoguernn,ghosal2019dialoguegcn}. 
ERC has emerged as an important research direction due to its broad applications in empathetic dialogue systems \citep{rashkin2019towards,zhou2018emotional}, mental health monitoring \citep{cummins2019review}, and understanding of dynamic human emotions.
This deeper comprehension has
meaningful impacts, such as supporting mental health monitoring and improving human–machine
interaction. As people increasingly interact with machines in their daily lives, it has become natural to expect these systems not only to understand the content of what is said, but also to recognize the emotions conveyed and to respond in ways that align with those emotional cues.

\begin{figure}[t]
\centering
\includegraphics[width=0.85\linewidth]{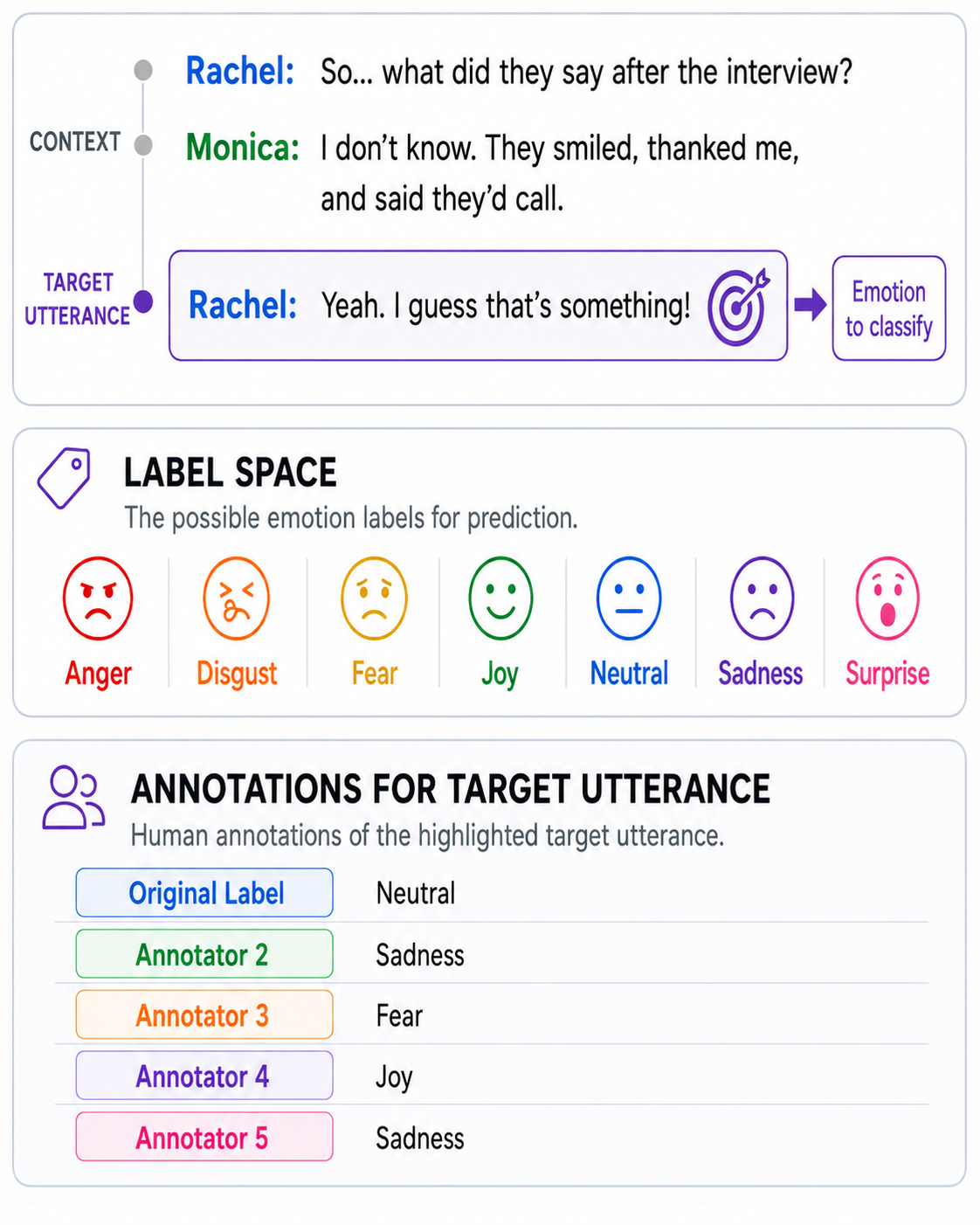}
\caption{Example of annotation ambiguity in ERC from MELD, where emotional interpretation depends on implicit intention and allows multiple valid labels.}

\label{fig:annotation_ambiguity_example}
\end{figure}

Research in ERC has been shaped by datasets such as MELD, EmoryNLP, DailyDialog,
and IEMOCAP \citep{poria2019meld,zahiri2018emorynlp,li2017dailydialog,busso2008iemocap}. Classical methods (e.g., DialogueRNN, DialogueGCN, COSMIC)
modeled speaker states and conversational structures \citep{majumder2019dialoguernn,ghosal2019dialoguegcn,ghosal2020cosmic}.
More recently, Large Language Models (LLMs) have been adopted, typically via fine-tuning, zero-shot or few-shot
prompting \citep{lei2023instructerc,shen2025coe}. Fine-tuned models tend to perform well within their training
domain when provided with auxiliary information such as scene descriptions or persona details, but they
struggle to generalize to datasets where such contextual signals are absent \citep{shen2025coe,fu-etal-2025-laerc,he-etal-2025-dialoguemmt}.

Despite these advances, a fundamental assumption underlying most ERC 
research has gone largely unquestioned: that each utterance in a 
conversation can be assigned a single, unambiguous correct emotion label. 
In practice, emotional expressions in dialogue are shaped 
by context, speaker intent, and subtle linguistic cues, making strict 
single-label annotation insufficient in many cases, like in the figure\ref{fig:annotation_ambiguity_example}. 
%This raises an 
% important question: when LLMs appear to predict a different emotion than the gold label on ERC benchmarks, do 
% these apparent errors necessarily reflect poor emotion understanding, or are 
% they, to a significant extent, a consequence of the single-label 
% paradigm itself,  one that does not account for the fact that multiple 
% emotions may be plausible for a given utterance?
This raises an important question: when LLMs predict an emotion that differs from the gold label in ERC benchmarks, do such discrepancies necessarily indicate poor emotion understanding, or do they partly reflect the limitations of the single-label paradigm, which fails to capture the possibility that multiple emotions may plausibly characterize the same utterance?

To investigate this question, we conduct a two-stage study. We first conduct a systematic zero-shot evaluation of LLMs on multiple ERC
benchmarks using their original annotations, revealing that while aggregate metrics suggest reasonable performance, 
minority emotions such as fear and disgust are 
systematically misclassified across all models. Crucially, these 
failures concentrate around utterances with specific linguistic 
properties: negations, exclamations, interjections, and short 
replies. Closer human inspection of these cases revealed that multiple emotional interpretations could plausibly be detected for the same utterance, suggesting that the errors are rooted in the data itself rather than model-specific weaknesses. 

To explore this further, we re-annotate representative subsets of these datasets with additional human annotators and analyze inter-annotator agreement, treating the original dataset label as an additional annotation. This distinguishes likely errors from cases where low human agreement suggests multiple plausible emotions.

In many existing ERC datasets, our findings suggest that ambiguity allows the same instance to reasonably support different emotion labels. 
%To enable fair evaluation and comparison at scale, this can be addressed either by extending the annotation of existing datasets with multiple emotion labels to capture such uncertainty. And also can be done by using an automatic evaluator (in our case is LLM as a judge) which tests whether a candidate label is valid given an utterance and its context. We pursue this by given a label, an utterance, and dialogue context, predicts label validity; comparison with human annotations yields strong results across all benchmarks, demonstrating its effectiveness as a reliable evaluation framework for emotion plausibility.
To enable fair and scalable evaluation, this issue can be addressed in two ways: by extending existing datasets with multiple plausible emotion labels to better capture annotation uncertainty, or by using an automatic evaluator that determines whether a candidate label is valid given an utterance and its dialogue context. We follow the latter approach and introduce an LLM-based evaluator that, given an emotion label, an utterance, and the surrounding dialogue, predicts whether the label is plausible. Its strong agreement with human annotations across all benchmarks demonstrates its effectiveness as a reliable framework for evaluating emotion-label plausibility.

Our main contributions are summarized as follows:
\begin{itemize}
    \item We present a systematic zero-shot assessment of multiple LLMs across MELD, EmoryNLP, and
    DailyDialog using full conversational context measuring performance against the original single label annotations.
    \item We re-annotate representative subsets of these datasets and analyse inter-annotater agreement, showing that strong consensus is limited across a substantial portion of ERC instances.
    \item We evaluate how well LLMs predict emotions when human annotations are treated as the reference labels, and analyze how performance varies across different levels
of inter-annotator agreement.

    \item We introduce an LLM-as-Judge framework as an evaluation approach that moves beyond the single label annotation constraints, achieving strong agreement with human annotations across all benchmarks.
\end{itemize}

\section{Related Work}

Emotion Recognition in Conversations has evolved from early feature-based
and neural approaches to more recent LLM-based methods.
Initial work focused on modeling temporal dynamics and inter-speaker interactions
using recurrent and graph-based architectures
\citep{majumder2019dialoguernn,shen-etal-2021-directed}, while later studies
incorporated external knowledge sources, such as commonsense reasoning features, to better capture contextual and affective dependencies
in dialogue \citep{zhong2019knowledge,ghosal2020cosmic}.
Despite these methodological advances, ERC benchmarks remain challenging due to
intrinsic dataset properties.
Most datasets are highly imbalanced, with neutral emotions dominating and
minority emotions sparsely represented \citep{poria2019meld,kang-cho-2024-improving}.
Moreover, emotion recognition is typically framed as a single-label
classification task, implicitly assuming a unique and unambiguous gold label for
each utterance.

Recent LLM-based approaches have demonstrated promising performance on ERC tasks.
Methods such as CoE: A Clue of Emotion Framework for Emotion Recognition in Conversations \citep{shen2025coe} leverage auxiliary information, including
scene descriptions and persona cues, to improve accuracy.
However, these gains often rely on dataset-specific contextual signals and do not
generalize well across benchmarks \citep{peng-etal-2025-emotion}.
In parallel, the emergence of large-scale LLMs such as LLaMA
\citep{touvron2023llama2}, Mistral \citep{jiang2023mistral}, Qwen \citep{bai2023qwentechnicalreport},
Gemma \citep{gemmateam2024gemmaopenmodelsbased}, and GPT-OSS \citep{openai2025gptoss} has reshaped ERC
research, with many studies relying on fine-tuning tailored
to specific benchmarks \citep{feng2024affect}.
However, comparatively little attention has been paid to zero-shot ERC evaluation
under full conversational context, where models are not adapted to
dataset-specific annotations \citep{10888198}.
Moreover, prior evaluations have paid limited attention to inter-annotator agreement, making it difficult to determine whether apparent model errors reflect genuine model limitations or ambiguity in the emotion labels themselves.
More importantly, existing ERC benchmarks and evaluation protocols rarely
consider the possibility that an utterance may admit multiple plausible emotional
interpretations, instead enforcing a single-label assumption that overlooks the
subjective and context-dependent nature of emotion perception.
\section{Experimental Setup}
Our study is conducted in two distinct stages. First, we perform a systematic zero-shot evaluation of LLMs across multiple ERC benchmarks using their original annotations. Second, we re-annotate representative subsets of these datasets with additional human annotators and analyze inter-annotator agreement, considering the original dataset label as an additional annotation.
\subsection{Datasets}
We evaluate four widely used ERC benchmarks that differ significantly in genre and class distribution: MELD \citep{poria2019meld}, EmoryNLP \citep{zahiri2018emorynlp}, DailyDialog \citep{li2017dailydialog}, and IEMOCAP \citep{busso2008iemocap}. MELD and EmoryNLP are both derived from the \textit{Friends} TV series; however, MELD uses seven standard emotions and is $\sim$48\% neutral, while EmoryNLP also uses seven emotions but employs a finer-grained, context-dependent inventory. DailyDialog consists of everyday two-speaker exchanges and exhibits a severe class imbalance, with over 80\% of utterances labeled as neutral. Finally, IEMOCAP contains scripted and improvised dyadic dialogues, from which they utilize six categorical labels to capture both acted and spontaneous emotional expressions. Table~\ref{tab:dataset_stats_erc} summarizes the main characteristics of the datasets used in this study, including their train, validation, and test splits, number of emotion classes, proportion of neutral utterances, and degree of class imbalance.
\begin{table}[t]
\centering
\setlength{\tabcolsep}{4pt}
\renewcommand{\arraystretch}{1.15}
\scalebox{0.9}{
\tiny
\begin{tabular}{l|ccc|ccc|c|c|c}
\hline
\textbf{Dataset}
& \multicolumn{3}{c|}{\textbf{Conversations}}
& \multicolumn{3}{c|}{\textbf{Utterances}}
& $\boldsymbol{|\mathcal{E}|}$
& \textbf{Neutral (\%)}
& \textbf{Imbalance Ratio} \\

& \textbf{Train}
& \textbf{Val}
& \textbf{Test}
& \textbf{Train}
& \textbf{Val}
& \textbf{Test}
& & & \\
\hline

\textbf{IEMOCAP}
& 100 & 20 & 31
& 4810 & 1000 & 1623
& 6 & 22.98 & 3:1 \\

\textbf{EmoryNLP}
& 713 & 99 & 85
& 9934 & 1344 & 1328
& 7 & 29.95 & 4:1 \\

\textbf{MELD}
& 1038 & 114 & 280
& 9989 & 1109 & 2610
& 7 & 48.21 & 18:1 \\

\textbf{DailyDialog}
& 11118 & 1000 & 1000
& 87170 & 8069 & 7740
& 7 & 83.24 & 1156:1 \\

\hline
\end{tabular}
}
\caption{Dataset statistics for the ERC benchmarks used in this study.
$\mathcal{E}$ denotes the number of emotion classes.
The neutral proportion and imbalance ratio illustrate the class
distribution skew in each dataset.}
\label{tab:dataset_stats_erc}
\end{table}

\subsection{Prompting Protocol}

We evaluate each target utterance using a contextual prompting setup, 
where all preceding dialog turns with speaker attribution are included to preserve conversational history. 
This follows prior ERC work showing that dialog context and speaker states improve emotion recognition 
\citep{majumder2019dialoguernn,ghosal2019dialoguegcn} 
as well as recent LLM-based studies highlighting the benefits of incorporating conversational history \citep{zhang2024dialoguellmcontextemotionknowledgetuned,hong2025aerllmambiguityawareemotionrecognition}.
We therefore adopt the full-context setup to assess each model’s intrinsic contextual reasoning ability. 
Decoding is performed with a low temperature of 0.3 to ensure consistency and comparability across models. 
\subsection{Evaluation Metrics}
We report five complementary metrics in the first stage of our systematic zero-shot evaluation of LLMs. Accuracy provides an overall measure of correctness but can be dominated by frequent classes in highly imbalanced datasets. Weighted F1 (W-F1) reflects overall performance while accounting for class imbalance, but may still obscure errors on minority emotions. Macro F1 (M-F1) assigns equal weight to all classes, making it essential in ERC where rare emotions often carry crucial affective signals. 
In the second stage of our evaluation, which focuses on human re-annotation, Cohen's $\kappa$ measures inter-annotator agreement beyond chance, accounting for the possibility that annotators may agree randomly; it is reported to contextualize model performance relative to human consistency on the same data. Finally, the False Positive Rate (FPR) captures how often a model predicts an emotion that is not supported by any human annotator, providing a direct measure of over-prediction that accuracy and F1 scores alone may miss.

\section{Zero-Shot ERC Performance and Disagreement Analysis}

In this section, we present zero-shot results of
four LLMs (LLaMA-70B, Qwen-32B, GPT-3.5, 
Mistral-7B) on three ERC benchmarks (MELD, 
EmoryNLP, DailyDialog). We look specifically at where models fail and whether those failures 
follow systematic patterns. 

\subsection{LLM performance Analysis}

\begin{table}[t]
\scriptsize
\centering
\renewcommand{\arraystretch}{1.1}
\setlength{\tabcolsep}{3pt}
\begin{tabular}{lcccc}
\hline
\textbf{Emotion / Metric} & \textbf{LLaMA-70B} & \textbf{Qwen-32B} & \textbf{GPT-OSS (120B)} & \textbf{Mistral-7B} \\
\hline
\multicolumn{5}{c}{\textbf{MELD}} \\
Anger & 0.556 & 0.528 & 0.542 & 0.429 \\
Disgust & 0.398 & 0.389 & 0.368 & 0.244 \\
Fear & 0.333 & 0.268 & 0.303 & 0.250 \\
Joy & 0.598 & 0.000 & 0.572 & 0.544 \\
Neutral & 0.722 & 0.716 & 0.700 & 0.708 \\
Sadness & 0.473 & 0.471 & 0.449 & 0.398 \\
Surprise & 0.560 & 0.577 & 0.539 & 0.373 \\
\textbf{Macro-F1} & \textbf{0.520} & 0.369 & 0.434 & 0.421 \\
\textbf{Weighted-F1} & \textbf{0.628} & 0.529 & 0.606 & 0.564 \\
\hline
\multicolumn{5}{c}{\textbf{EmoryNLP}} \\
Joyful & 0.538 & 0.524 & 0.532 & 0.464 \\
Mad & 0.391 & 0.438 & 0.408 & 0.400 \\
Neutral & 0.493 & 0.501 & 0.527 & 0.503 \\
Peaceful & 0.031 & 0.099 & 0.062 & 0.105 \\
Powerful & 0.110 & 0.114 & 0.097 & 0.048 \\
Sad & 0.362 & 0.392 & 0.335 & 0.332 \\
Scared & 0.264 & 0.403 & 0.320 & 0.062 \\
\textbf{Macro-F1} & 0.274 & \textbf{0.309} & 0.228 & 0.239 \\
\textbf{Weighted-F1} & 0.355 & \textbf{0.388} & 0.372 & 0.315 \\
\hline
\multicolumn{5}{c}{\textbf{DailyDialog}} \\
Anger & 0.420 & 0.479 & 0.359 & 0.430 \\
Disgust & 0.226 & 0.261 & 0.284 & 0.201 \\
Fear & 0.158 & 0.129 & 0.121 & 0.117 \\
Happiness & 0.538 & 0.560 & 0.551 & 0.455 \\
Neutral & 0.807 & 0.829 & 0.778 & 0.685 \\
Sadness & 0.249 & 0.244 & 0.272 & 0.146 \\
Surprise & 0.375 & 0.367 & 0.306 & 0.225 \\
\textbf{Macro-F1} & 0.396 & \textbf{0.410} & 0.382 & 0.323 \\
\textbf{Weighted-F1} & 0.746 & \textbf{0.769} & 0.724 & 0.633 \\
\hline
\end{tabular}
\caption{Per-emotion F1 scores and aggregate Macro-F1 and Weighted-F1 scores across datasets and models. Bold values indicate the highest aggregate score among the evaluated models for each dataset and metric.}
\label{tab:final_all_models_emotions}
\end{table}

\begin{figure}[t]
\centering
\includegraphics[width=1\linewidth]{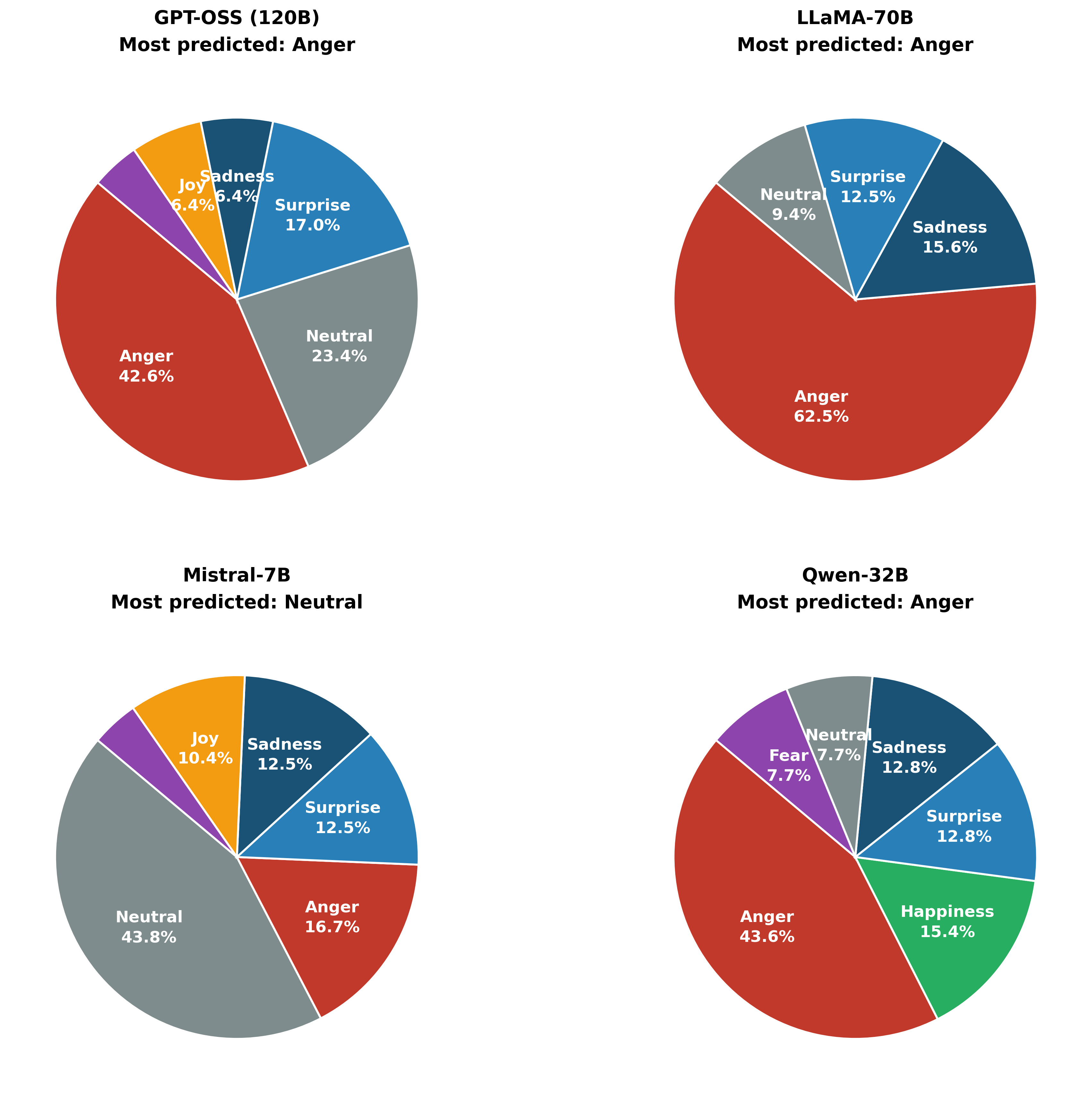}
\caption{Predicted emotion distributions for \textbf{Disgust} utterances in MELD across all models.}
\label{fig:meld_pies}
\end{figure}

Table~\ref{tab:final_all_models_emotions} reports per-emotion F1 scores across all models and datasets. While Weighted-F1 scores appear relatively high on MELD and DailyDialog (0.60–0.75), Table~\ref{tab:final_all_models_emotions} reveals a more troubling reality: models consistently struggle with minority emotions across all benchmarks. As shown in Figure~\ref{fig:meld_pies}, Disgust utterances in MELD are frequently misclassified as Anger or Neutral. Weighted-F1 drops sharply for EmoryNLP (0.31–0.39), while Macro-F1 remains low across all datasets, confirming that rare emotions such as Fear and Disgust are not reliably recognized in zero-shot settings.

Dataset-level interpretation suggests that MELD and DailyDialog contain clearer and more repetitive emotional categories, making them easier for models to classify. In contrast, EmoryNLP includes subtle, context-dependent emotions like Powerful and Peaceful, which are inherently more difficult to detect (Table~\ref{tab:final_all_models_emotions}).

Despite these dataset-specific differences in overall performance and label structure, the same conclusions hold across all benchmarks: models perform better on dominant emotion classes, struggle with rare or ambiguous categories, and obtain Weighted-F1 scores that mask substantial weaknesses at the class level.

\subsection{Consistent Failure Modes in Utterances}

To identify recurring linguistic patterns associated with misclassifications under the original single-label annotations, we first manually inspected 300 misclassified utterances (100 per dataset).  
This initial review revealed four dominant sources of confusion: negations, exclamations, interjections, and short or minimal replies.  
We then extended this analysis automatically across the entire test sets using a simple rule-based tagging system.  
For each utterance, we automatically detected whether it contained a negation word (e.g., “not”, “don’t”), an exclamation mark, a common interjection (e.g., “ugh”, “oh”, “hmm”), or a question mark, and whether it was shorter than five words.  
By cross-referencing these tags with model predictions, we computed how often each linguistic pattern coincided with a misclassification with respect to the original single label ground truth. The code and all accompanying materials are provided in the supplementary materials.
\begin{figure}[t]
    \centering
    \includegraphics[width=\linewidth]
    {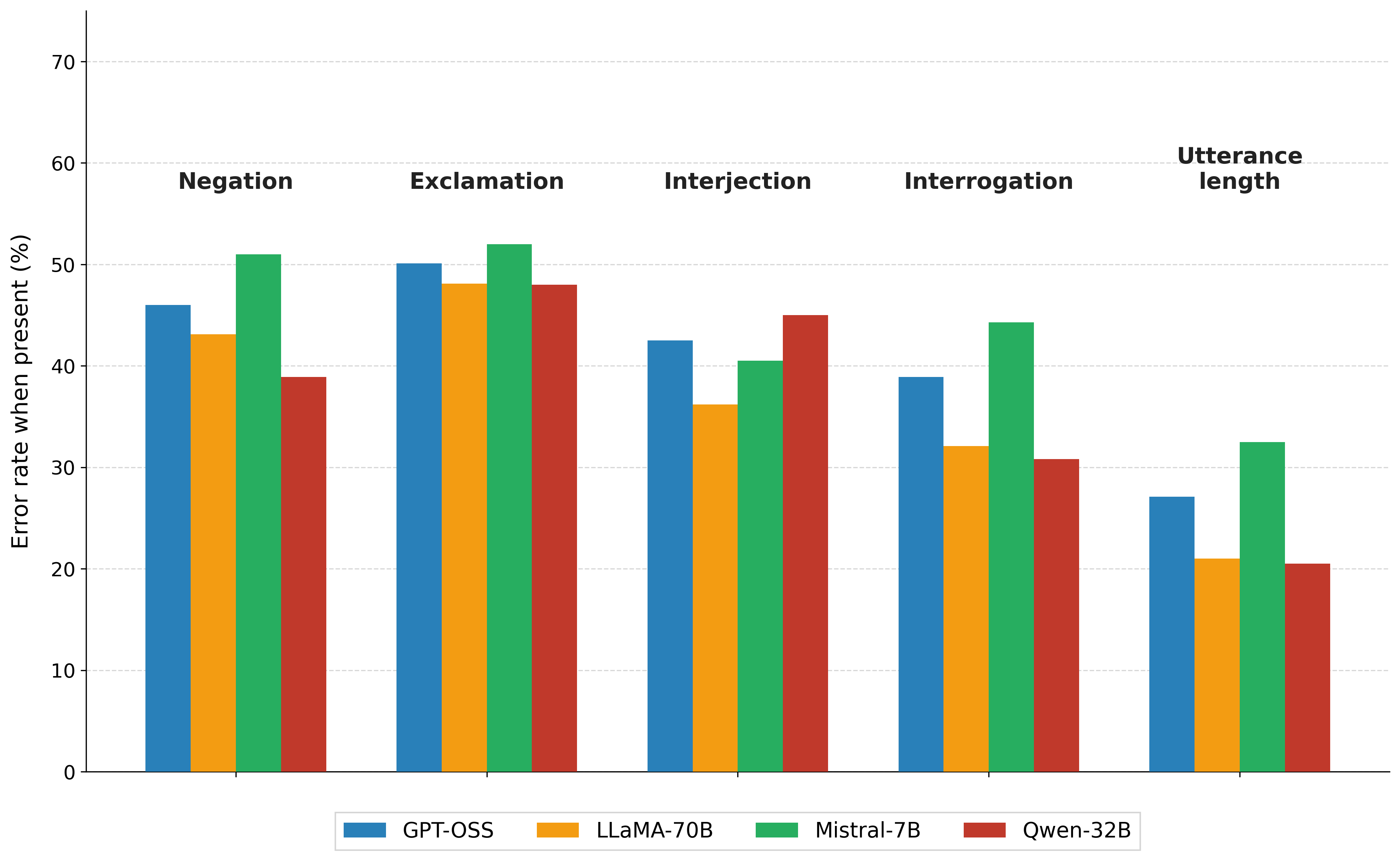}
    \caption{Comparison of failure-mode error rates in DailyDialog across four LLMs
    (GPT-OSS-120B, LLaMA-70B, Mistral-7B, and Qwen-32B).
    Negations and exclamations show the highest error rates, followed by
    interjections and interrogations, while short replies result in fewer errors.}
    \label{fig:failure_modes}
\end{figure}

Figure~\ref{fig:failure_modes} summarizes the results for DailyDialog.
Across all models, negation and exclamation consistently produce the highest error rates (around 45–50\%), followed by interjections and interrogations, while short utterances cause fewer errors. The supplementary materials report equivalent analyses for MELD, EmoryNLP and DailyDialog showing that models encounter similar linguistic failure modes, including negations, exclamations.

However, manual inspection of a subset of these cases indicates that multiple emotional interpretations may be plausible given the context, the intent, the existing lexical cues and punctuations.The consistency of these failure patterns across models and datasets raises a 
question that aggregate metrics cannot answer: are these genuine 
model limitations, or do they reflect something more fundamental 
about the data itself? This question motivates an analysis of human annotation agreement in ERC benchmarks. 
\section{Human Agreement and Label Ambiguity in ERC Benchmarks}
To better understand whether the misclassification errors observed in the previous section stem from genuine model limitations or from ambiguity 
in 
utterances and their context, we run a small-scale annotation task of emotions on the different datasets and 
we analyze human
agreement. 
Specifically, we investigate how often annotators agree versus diverge in emotion labeling and assess the extent to which ERC benchmarks exhibit intrinsic ambiguity, potentially accounting for these 
differences.

\subsection{Re-Annotation Setup and Agreement Measurement}
\begin{figure}[t]
    \centering
    \includegraphics[scale = 0.25]{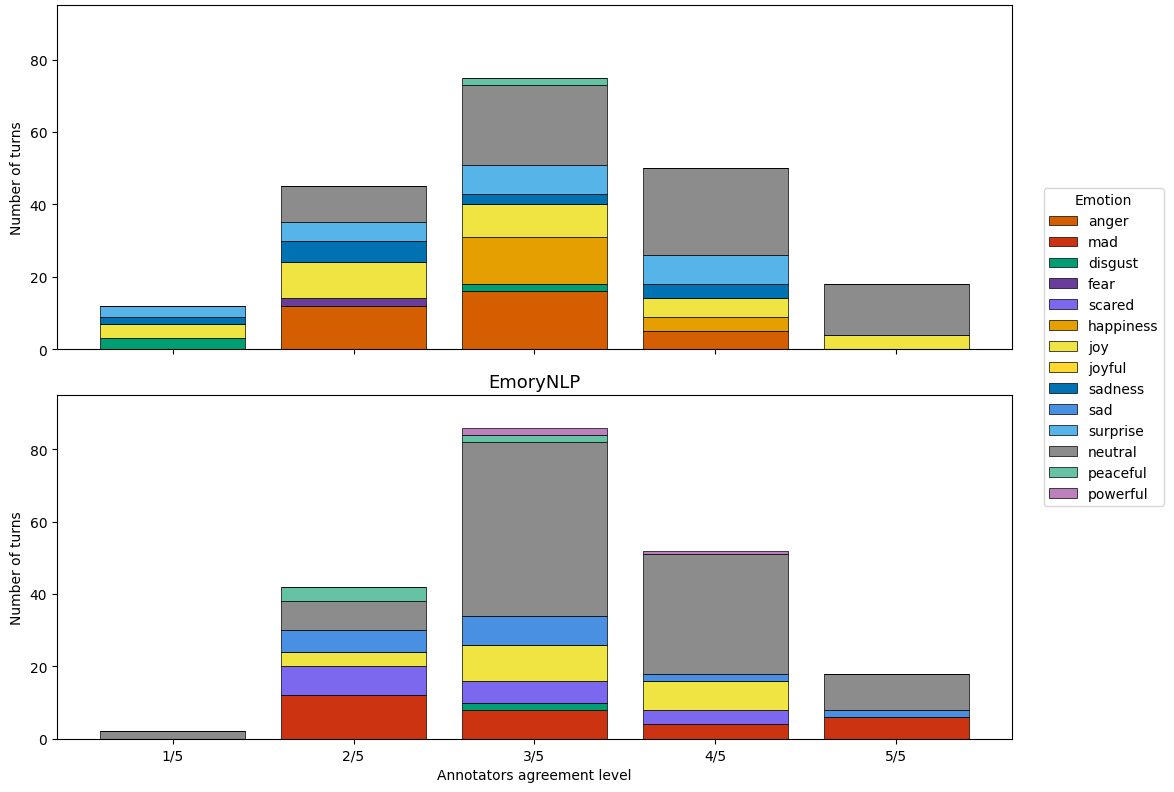}
    \caption{Annotator agreement distributions for MELD (top) and EmoryNLP (down), stratified by emotion category. High-consensus cases (4/5 and 5/5) are dominated by the neutral class, while minority emotions concentrate in intermediate agreement regimes (2/5 and 3/5).}
    \label{fig:agreement-meld-emory}
\end{figure}

Since ERC benchmarks rely on a single emotion label per utterance, we conducted a controlled re-annotation study on representative subsets of each dataset (All annotation materials, including the full protocol, annotator instructions, and emotion label definitions, are provided in the supplementary materials to support transparency and reproducibility).
For MELD, EmoryNLP, DailyDialog, and IEMOCAP, we sampled 200 utterances drawn from different conversations to ensure a diverse range of conversational contexts. The sampling process was designed to ensure that all emotion categories present in each dataset were included, allowing the annotation agreement to be examined across the entire label space.
Each selected turn was re-annotated by four 
human annotators who 
were instructed to assign a \emph{single emotion label} chosen from the original emotion set of the corresponding dataset.
The original dataset annotation was treated as an additional annotator, resulting in five annotations per turn. Inter-annotator agreement was assessed using Cohen's $\kappa$, computed pairwise between all possible annotator pairs, and averaged across pairs per dataset. 
The resulting agreement scores are consistently low across the benchmarks, with $\kappa=0.43$ for MELD with the help of multimodal information, $0.14$ for EmoryNLP, $0.29$ for DailyDialog and $0.27$ for IEMOCAP.
These values are comparable to, and in some cases lower than, those reported in
the original datasets' papers \citep{poria2019meld,zahiri2018emorynlp,li2017dailydialog,busso2008iemocap},
providing further evidence that low annotator agreement reflects a broader property of ERC data: many utterances admit multiple plausible emotional interpretations.
Beyond that, we analyze agreement at the instance level by grouping turns according to the number of annotators assigning the same emotion.
We define five agreement regimes: \textit{5/5} (full agreement), \textit{4/5} (strong agreement), \textit{3/5} (moderate agreement), \textit{2/5} (low agreement), and \textit{1/5} (complete disagreement).
Figure~\ref{fig:agreement-meld-emory} presents agreement histograms for MELD and EmoryNLP, with bars further decomposed by emotion category.

The histograms of Figure \ref{fig:agreement-meld-emory} reveal that the majority of annotated utterances fall into moderate (3/5) and low (2/5) agreement regimes, indicating that genuine ambiguity is the norm rather than the exception. High-consensus cases (4/5 and 5/5) account for only a minority of utterances and are largely dominated by  \textit{neutral} emotion.
In contrast, minority emotions such as \textit{fear}, \textit{disgust}, \textit{sadness},  \textit{peaceful}, and \textit{powerful} appear predominantly in the 1/5, 2/5 and 3/5 agreement regimes.
This pattern indicates that these emotions are not only underrepresented in the datasets, but also intrinsically more ambiguous for annotators.
Rather than reflecting random disagreement, this indicates that an utterance can support multiple plausible emotional
interpretations. Although Figure~\ref{fig:agreement-meld-emory} presents only MELD and EmoryNLP, the agreement distributions for all evaluated datasets are reported in the supplementary materials. The same overall conclusion is observed across all datasets: dominant and neutral categories tend to receive higher annotator agreement, whereas minority emotions are concentrated in lower-agreement regimes.

These findings highlight a structural limitation of single-label ERC benchmarks: many instances lack a clear emotional consensus even among humans, which make the LLM models hard to choose the right emotion. This directly questions the validity of single-label ground truth. 
\subsection{Linguistic and Contextual Factors Driving Annotation Disagreement}
\begin{table}[t]
\centering
% \scriptsize
\setlength{\tabcolsep}{2pt}
\begin{tabular}{|l|c|c|c|c|}
\hline
\textbf{Dataset} & \textbf{Acc.} & \textbf{Prec.} & \textbf{Rec.} & \textbf{F1} \\
\hline
MELD        & 0.84 & 0.78 & 0.84 & 0.80 \\
DailyDialog & 0.76 & 0.79 & 0.76 & 0.71 \\
EmoryNLP    & 0.58 & 0.70 & 0.58 & 0.61 \\
IEMOCAP     & 0.68 & 0.77 & 0.68 & 0.66 \\
\hline
\end{tabular}
\caption{Decision tree performance for predicting annotation agreement levels using linguistic and contextual cues. Agreement is modeled using three classes: low agreement (1/5), medium agreement (2/5--3/5), and high agreement (4/5--5/5). The decision tree is trained on 75\% of the annotated data and evaluated on the remaining 25\%.}
\label{tab:decision-tree-results}
\end{table}
To analyze the sources of annotation disagreement, we manually label 100 turns among 200 turns per dataset using five qualitative cues: \textit{Intent Clarity}, which distinguishes between clear and ambiguous communicative purposes; \textit{Intent Count}, identifying whether an utterance conveys a single intent or multiple ones (e.g., informing, requesting information, warning, encouraging, complaining, agreeing, mocking, or reproaching); \textit{Context Clarity}, assessing if the surrounding dialogue provides sufficient information to resolve the emotion; \textit{Lexical Cue Presence}, noting the existence of explicit affective terms; and \textit{Punctuation}, which categorizes markers like ``!!!'' or ``???'' that amplify intensity. Together, these cues allow us to identify how ambiguity arises from sparse context, overlapping intents, and the reliance on subjective inference in the absence of explicit lexical.
To assess whether these linguistic and contextual cues are predictive of annotator agreement levels, we train a decision tree classifier using the annotated cues as features and the agreement level as the target variable.

Table~\ref{tab:decision-tree-results} shows that linguistic and contextual cues help predict annotation agreement levels. The classifier performs best on MELD (0.84 accuracy) and DailyDialog (0.76), while results are lower on IEMOCAP (0.68) and EmoryNLP (0.58), consistent with their more subtle and context-dependent emotion inventories where ambiguity is harder to capture through surface-level cues alone. This suggests that annotation disagreement is not random, but can be explained by observable cues such as intent clarity, context, lexical emotion cues, and punctuation.

\begin{figure}[t]
    \centering
    \includegraphics[width=\linewidth]{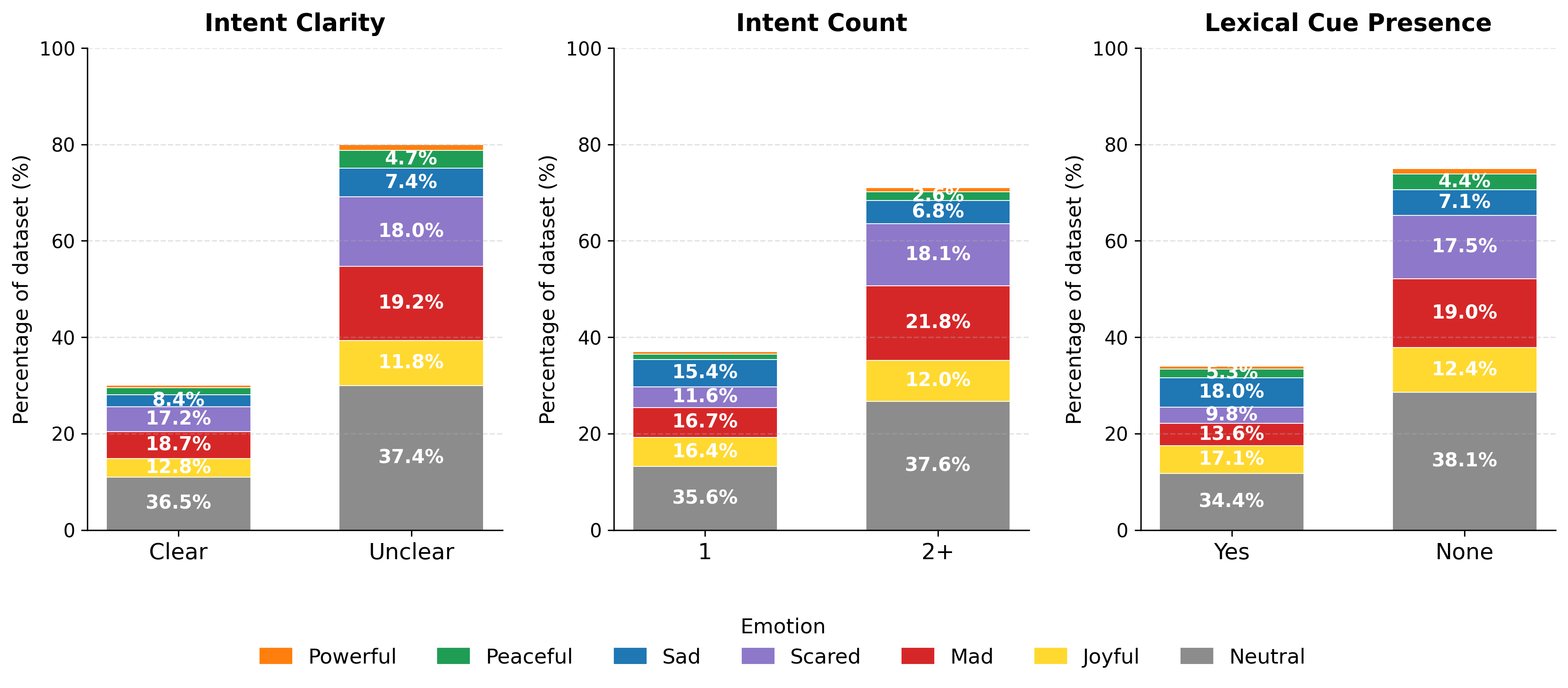}
    \caption{Distribution of linguistic and contextual cues (Intent Clarity, Intent Count, and Lexical Cue Presence) for EmoryNLP utterances with 2/5 and 3/5 annotator agreement. Ambiguous instances are predominantly associated with unclear intent, multiple competing intents, and absence of lexical emotion cues.}
    \label{fig:cue-distribution-23-a}
\end{figure}

\begin{figure}[t]
    \centering
    \includegraphics[width=0.9\linewidth]{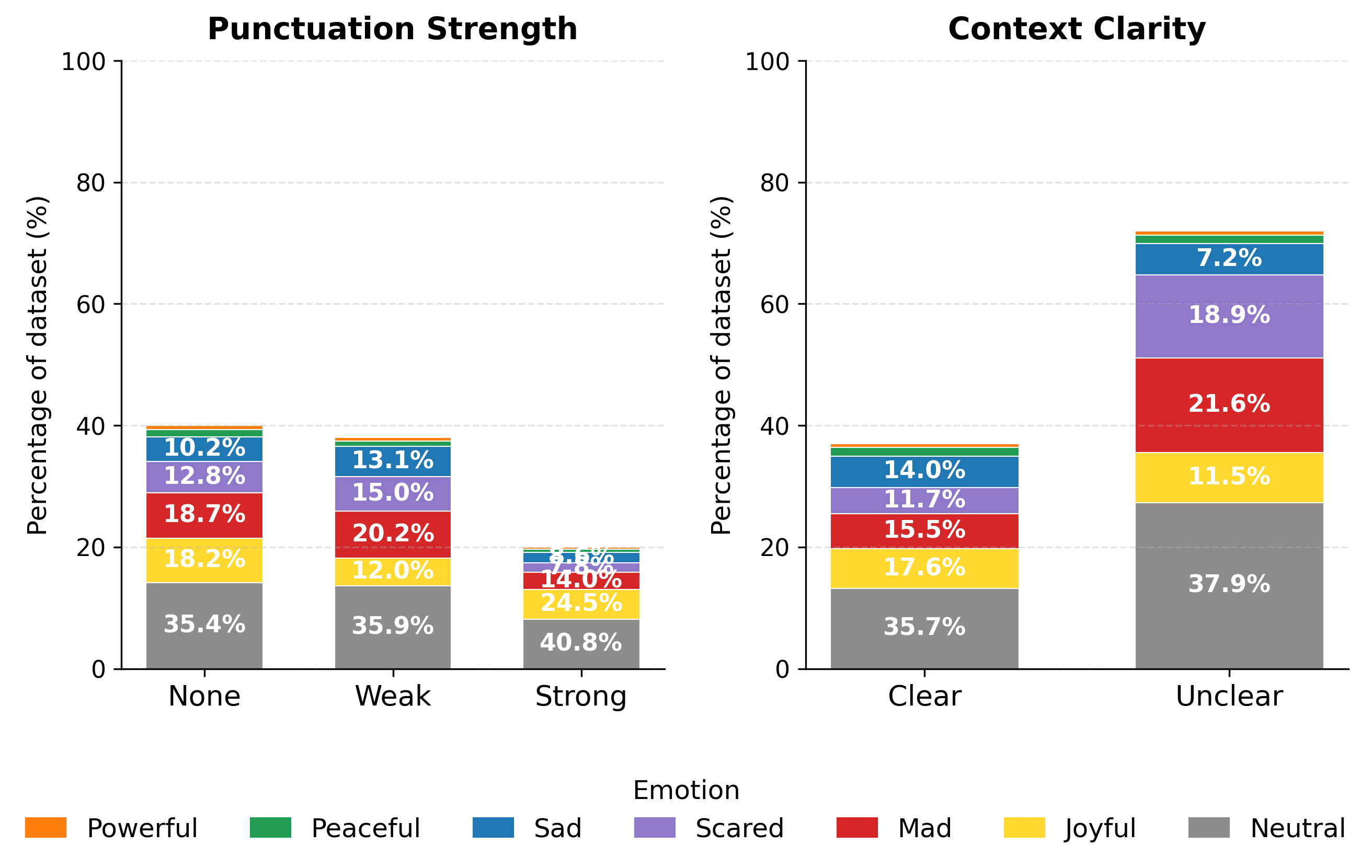}
    \caption{Distribution of linguistic and contextual cues (Punctuation Strength and Context Clarity) for EmoryNLP utterances with 2/5 and 3/5 annotator agreement. Ambiguous instances are predominantly associated with unclear context and weak or strong punctuation.}
    \label{fig:cue-distribution-23-b}
\end{figure}

To better understand the linguistic and contextual properties underlying these ambiguous cases, we focus specifically on turns in 2/5 and 3/5 agreement.
For these turns, we analyze the distribution of the manually annotated cues capturing the characteristics of the utterances.
Figure~\ref{fig:cue-distribution-23-a} and Figure ~\ref{fig:cue-distribution-23-b} presents the cue distributions for ambiguous instances.

Several consistent patterns emerge from this analysis. First, among 2/5 and 3/5 agreement utterances, unclear intent is the dominant category (Figure~\ref{fig:cue-distribution-23-a} and Figure~\ref{fig:cue-distribution-23-b}), suggesting that annotators struggle to infer a single communicative goal when the speaker's intent is ambiguous. Second, utterances with multiple competing intents (2+) are more frequent than single-intent utterances, confirming that overlapping communicative goals are a key driver of disagreement. Third, the absence of explicit lexical emotion cues is prevalent, forcing annotators to rely on contextual inference. Finally, unclear context is overrepresented across these cases, further increasing interpretive variability and disproportionately affecting minority emotion categories. By contrast, utterances in the 1/5 agreement regime exhibit the reverse pattern across these dimensions; with the corresponding analysis to be provided in the supplementary materials.

These findings have direct implications for ERC evaluation. If several utterances admit multiple plausible emotional interpretations, as reflected by human annotations, then penalizing models for predicting a plausible label different from the original single annotation may mischaracterize their performance.
This
motivates considering the evaluation of ERC models on existing benchmarks
beyond single-label, allowing for different emotion perspectives.
\section{LLM Agreement under Multi-Annotated Emotion Scenarios}

\begin{table*}[t]
\centering
\scriptsize
\setlength{\tabcolsep}{2pt}
\renewcommand{\arraystretch}{1.05}
\begin{tabular}{|l|c|c|c|c|c|c|c|c|c|c|}
\hline
\textbf{Dataset} &
\textbf{\shortstack{LLaMA \\ 70B}} &
\textbf{GPT-OSS} &
\textbf{\shortstack{Qwen \\ 32B}} &
\textbf{\shortstack{Mistral \\ 7B}} &
\textbf{\shortstack{Gemma \\ 27B}} &
\textbf{DeepSeek-R1} &
\textbf{\shortstack{LLaMA \\ 8B}} &
\textbf{\shortstack{GPT-OSS \\ Safe}} &
\textit{\shortstack{Inter-model \\ agreement on \\ predicted emotion (\%)}} &
\textit{\shortstack{Agreement on \\ neutral and \\ positive emotions (\%)}}
\\
\hline
MELD        & 0.84 & 0.86 & 0.91 & 0.89 & 0.93 & 0.90 & 0.79 & 0.92 & 69 & 83 \\
DailyDialog & 0.91 & 0.96 & 0.97 & 0.88 & 0.89 & 0.95 & 0.81 & 0.93 & 65 & 86 \\
EmoryNLP    & 0.80 & 0.85 & 0.91 & 0.86 & 0.93 & 0.81 & 0.74 & 0.88 & 66 & 80 \\
IEMOCAP     & 0.82 & 0.76 & 0.83 & 0.70 & 0.88 & 0.81 & 0.75 & 0.85 & 59 & 76 \\
\hline
\end{tabular}
\caption{Multi-annotation--aware accuracy of eight LLMs across ERC benchmarks.
A prediction is considered correct if it matches at least one human-annotated emotion.
Models exhibit strong inter-model agreement, largely driven by neutral and positive emotions,
while minority emotions remain challenging across datasets.}
\label{tab:llm-multi-annotation}
\end{table*}
The findings in the previous section show that many ERC utterances do not have a clear agreed emotion label. So this raised a point: If multiple emotions are valid for the same utterance, then penalizing a model for not matching the original single label is unfair. Thus, we re-evaluate LLM predictions under a multi-annotation-aware setting, \textbf{where a prediction is considered correct if it matches at least one of the human-annotated emotions}.
We evaluate the eight LLMs listed in Table~\ref{tab:llm-multi-annotation}, which vary in size and architectural properties, using a zero-shot prompt with full conversational context across ERC benchmarks. Table~\ref{tab:llm-multi-annotation} reports the resulting accuracies, while the complete prompt template and evaluation procedure are provided in the supplementary materials.
Across datasets, we observe a convergence in model predictions, with inter-model agreement on a particular emotion among the emotion set, reaching 69\% on MELD, 65\% on DailyDialog, 59\% on IEMOCAP, and 66\% on EmoryNLP.
This convergence indicates that, despite architectural diversity, LLMs tend to favor similar emotional interpretations when multiple plausible labels are allowed.
However, agreement is strongly emotion-dependent.
Across all datasets, more than 80\% of agreed predictions correspond to \textit{neutral} or positive emotions such as \textit{joy} whereas minority emotions including \textit{disgust} and \textit{fear} remain consistently miss-predicted mostly by majorty-class emotions.
This pattern mirrors the human annotation distributions observed in the Section of Linguistic and Contextual Factors Driving
Annotation Disagreemet.
Overall, these findings motivate evaluation beyond strict single-label emotion.

\section{Evaluation with LLM-as-Judge}
\begin{table}[t]
\centering
\setlength{\tabcolsep}{4pt}
\scriptsize
\renewcommand{\arraystretch}{1}
{
\begin{tabular}{|l|l|c|c|c|c|c|}
\hline
\textbf{Dataset} &
\textbf{Judge} &
\textbf{Kappa} &
\textbf{Comp.} &
\textbf{Acc.} &
\textbf{Rec.} &
\textbf{FPR} \\
\hline
DailyDialog & Gemma-3-27B  & 0.60 & 0.91 & 0.86 & 0.84 & 0.14 \\
& LLaMA-70B     & 0.67 & 0.88 & 0.85 & 0.83 & 0.15 \\
& Qwen2.5-32B   & 0.57 & 0.69 & 0.79 & 0.76 & 0.21 \\
& Gemma-4-31B  & 0.66 & 0.82 & 0.84 & 0.82 & 0.16 \\
& Gemma-4-26B   & 0.62 & 0.79 & 0.83 & 0.81 & 0.17 \\
& Qwen3.5-35B  & 0.65 & 0.84 & 0.85 & 0.83 & 0.15 \\
\hline
MELD & Gemma-3-27B   & 0.47 & 0.90 & 0.82 & 0.80 & 0.18 \\
& LLaMA-70B     & 0.55 & 0.80 & 0.81 & 0.79 & 0.19 \\
& Qwen2.5-32B   & 0.40 & 0.50 & 0.72 & 0.70 & 0.28 \\
& Gemma-4-31B   & 0.54 & 0.72 & 0.80 & 0.78 & 0.20 \\
& Gemma-4-26B   & 0.51 & 0.70 & 0.79 & 0.77 & 0.21 \\
& Qwen3.5-35B   & 0.56 & 0.76 & 0.81 & 0.79 & 0.19 \\
\hline
IEMOCAP & Gemma-3-27B   & 0.32 & 0.90 & 0.74 & 0.72 & 0.26 \\
& LLaMA-70B     & 0.49 & 0.86 & 0.76 & 0.74 & 0.24 \\
& Qwen2.5-32B   & 0.36 & 0.50 & 0.68 & 0.66 & 0.32 \\
& Gemma-4-31B   & 0.46 & 0.81 & 0.74 & 0.72 & 0.26 \\
& Gemma-4-26B  & 0.44 & 0.78 & 0.73 & 0.71 & 0.27 \\
& Qwen3.5-35B   & 0.47 & 0.79 & 0.74 & 0.72 & 0.26 \\
\hline
EmoryNLP & Gemma-3-27B   & 0.43 & 0.86 & 0.78 & 0.76 & 0.22 \\
& LLaMA-70B     & 0.53 & 0.82 & 0.79 & 0.77 & 0.21 \\
& Qwen2.5-32B   & 0.43 & 0.45 & 0.70 & 0.68 & 0.30 \\
& Gemma-4-31B   & 0.52 & 0.69 & 0.77 & 0.75 & 0.23 \\
& Gemma-4-26B   & 0.49 & 0.67 & 0.76 & 0.74 & 0.24 \\
& Qwen3.5-35B   & 0.51 & 0.74 & 0.78 & 0.76 & 0.22 \\
\hline
\end{tabular}
}
\caption{LLM-as-Judge results across ERC datasets. Models are compared within each dataset.}
\label{tab:llm-judge-results}
\end{table}
The findings in the previous sections point to a common issue: many ERC utterances can reasonably support more than one emotional 
interpretation, yet current evaluation protocols penalize any 
prediction that does not match the original gold label. To address this, we introduce an \emph{LLM-as-Judge}
framework that \textbf{evaluates the plausibility of a given candidate emotion, given a target utterance and its full conversational context}. The judge is not asked to predict an emotion. Instead, giving the utterance and its context, presenting one emotion at a time and asking whether the emotion is plausible. This 
formulation naturally allows for multiple valid emotional 
interpretations of the same utterance, moving beyond the 
constraints of exact label matching.
We use different LLMs as judges to provide complementary reasoning behaviors across different model families and sizes. This allows us to examine whether the plausibility of emotion labels is judged consistently across heterogeneous LLM evaluators, rather than relying on a single model's interpretation of conversational context. Table~\ref{tab:llm-judge-results} reports the performance of all LLM judges across ERC datasets. Accuracy and recall measure how well these binary decisions align with human annotations at the emotion level. \textbf{The compatibility score measures whether the judge's plausibility decisions align with human annotations: it is positive for a given (utterance, emotion) pair when the judge correctly identifies an emotion as plausible and at least one human annotator assigned it, or when the judge correctly identifies an emotion as not plausible and no human annotator assigned it}. Cohen's kappa is reported as a measure of agreement between judge decisions and human annotations.

Overall, the results are strong and consistent across all judges and datasets. Compatibility scores range from 0.69 to 0.91 across all models and benchmarks, demonstrating that LLM judges reliably recognize emotions that human annotators considered valid. Accuracy ranges from 0.68 to 0.86, with DailyDialog consistently yielding the strongest performance across all judges, while IEMOCAP remains the challenging dataset. This pattern holds regardless of model family or size: larger models such as LLaMA-70B and Gemma-4-31B generally perform slightly better, but the differences are modest, suggesting that the difficulty is driven by the inherent ambiguity of the datasets rather than by model-specific capabilities or limitations. Cohen's kappa values follow the same trend. Importantly, False Positive Rates (FPR) remain low across all models and datasets, indicating that judges are conservative in their plausibility judgments and do not arbitrarily accept emotions as plausible. Taken together, these results demonstrate that LLM-based evaluation of emotion plausibility is both reliable and consistent across diverse model families. This suggests that they could provide a scalable and less costly complement to human reannotation when evaluating ambiguous ERC examples.

\section{Conclusion}
\label{sec:conclusion}

In this work, we presented a comprehensive analysis of ERC under a zero-shot setting, highlighting limitations of ERC
benchmarks and evaluation practices. By combining large-scale LLM evaluation,
re-annotation and agreement analysis, cue-based
investigation, and an LLM-as-Judge framework, we show that a substantial portion of ERC data is not well captured by strict single-label annotations. Our results show that many apparent model errors reflect genuine uncertainty. These findings position our LLM-as-Judge framework as strong for moving beyond strict labeling without requiring extensive human re-annotation.

\bibliography{aaai2026}

\end{document}